\documentclass{article}

\PassOptionsToPackage{numbers, compress}{natbib}

    \usepackage[preprint]{neurips_2025}

\usepackage{subcaption}

\usepackage[utf8]{inputenc} 
\usepackage[T1]{fontenc}    
\usepackage{hyperref}       
\usepackage{url}            
\usepackage{booktabs}       
\usepackage{amsfonts}       
\usepackage{nicefrac}       
\usepackage{microtype}      
\usepackage{xcolor}         
\usepackage{amsmath}    
 \usepackage{graphicx}    
\usepackage[normalem]{ulem}  

\usepackage[capitalize]{cleveref}
 \usepackage{multirow} 

\definecolor{ramenColor}{HTML}{0455cf}   
\definecolor{pizzaColor}{HTML}{D55E00}   

\definecolor{mygreen}{RGB}{4, 135, 76}
\definecolor{myblue}{RGB}{35, 80, 153}

\setcitestyle{numbers}

\title{SpeLLM: Input Tokens, Output Chars}
\title{SpeLLM: Character-Level Multi-Head Decoding}

\author{Amit Ben-Artzy \quad\quad Roy Schwartz \\
        The Hebrew University of Jerusalem \\
        \texttt{\{amit.benartzy, roy.schwartz1\}@mail.huji.ac.il}
        }

\begin{document}

\maketitle

\begin{abstract}

Scaling LLM vocabulary is often used to reduce input sequence length and alleviate attention's quadratic cost. Yet, current LLM architectures impose a critical bottleneck to this procedure: the output projection layer scales linearly with vocabulary size, rendering substantial expansion impractical.
We propose \textit{SpeLLM}, a method that decouples input and output vocabularies by predicting character-level strings through multiple output heads. In SpeLLM, each of the $k$ linear heads predicts a single character simultaneously, enabling the model to represent a much larger output space using smaller, independent linear heads. 
We present a self-distillation approach for converting a standard LLM to a SpeLLM. Our experiments with four pre-trained LLMs show their SpeLLM variants achieve competitive performance on downstream tasks while reducing runtime by $5.1\%$ on average across models. Our approach provides a potential avenue for reducing LLM costs, while increasing support for underrepresented languages and domains.\footnote{We release our code at \url{https://github.com/schwartz-lab-NLP/SpeLLM}}

\end{abstract}

\section{Introduction}
Large language models (LLMs) tend to use byte-pair encoding ~\citep{bpe1,bpe2} to both represent input text and generate the next token. Expanding the BPE vocabulary reduces sequence length, which in turn lowers both runtime and memory consumption.\footnote{Indeed, recent LLM vocabulary sizes reach 100k and beyond~\citep{llama_3,gemma_2}.}
Scaling this even further is particularly advantageous for non-English languages, where underrepresentation leads to longer token sequences and higher computational cost~\citep{ahia-etal-2023-languages,sengupta2023jais, petrov2024language}. 

However, it also raises the costs of generating each token, as the vocabulary embedding table is used both for projecting the output token at the final transformer layer, and computing the softmax probability vector for generating the next token.

Recent work addressed this tradeoff by training byte-level models from scratch~\cite{mega_byte,meta_blt}, which allow expanding the vocabulary without the accompanying costs. A different line of work has shown that standard LLMs~(trained using BPE) are capable of spelling~\citep{spelling_bee, what_do_tokens_know}. We therefore ask---can we adapt existing LLMs to generate their output tokens as sequences of characters rather than elements in a large embedding table?
 
\begin{figure}[!t]
  \centering
    \includegraphics[width=1\textwidth]{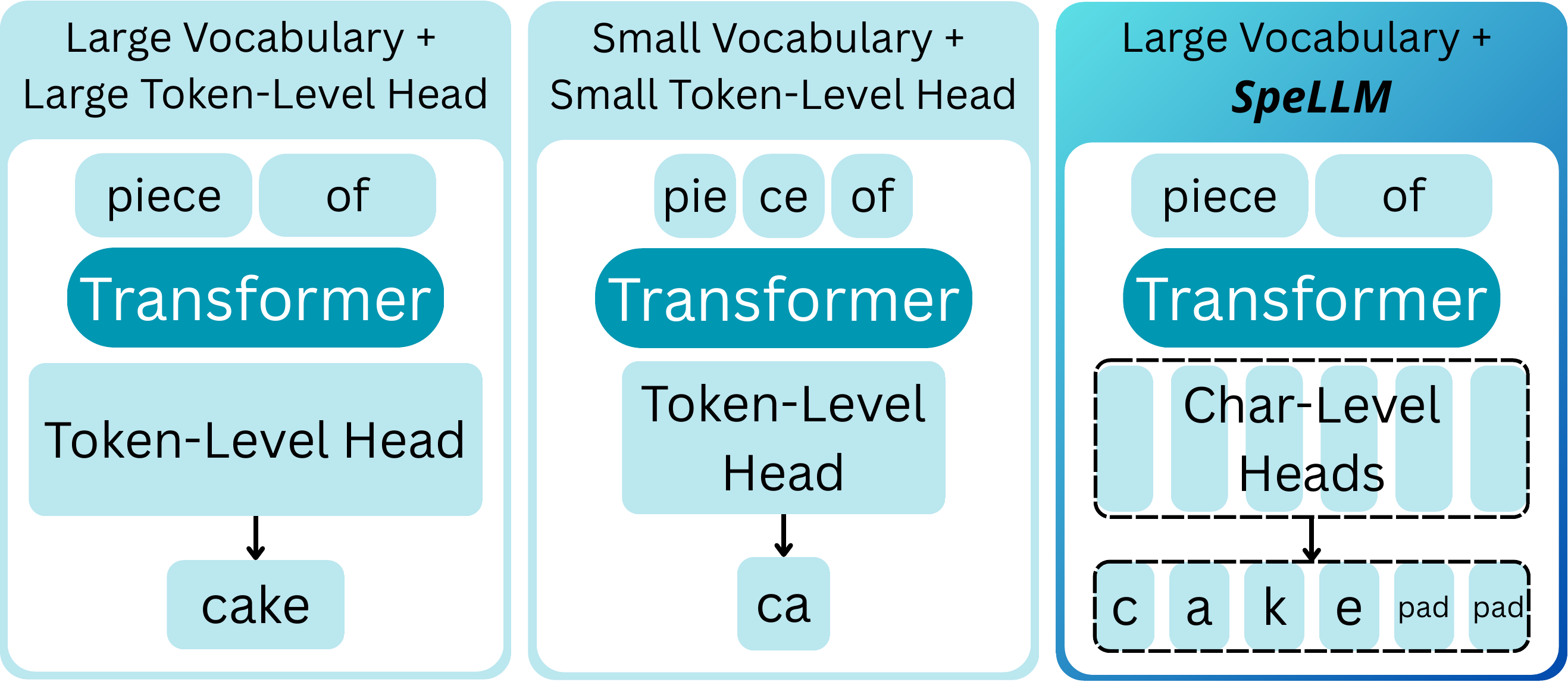}

  \caption{ 
\textbf{Left}: Using a large vocabulary allows us to represent the same text using fewer tokens, which reduces both space and runtime. However, it also requires projecting the output token onto a large embedding table, increasing computational cost and runtime of each decoding step;  \textbf{Middle}: Using a smaller vocabulary reduces the processing time of each token, but requires representing text using more tokens; \textbf{Right}:
  We propose \textit{SpeLLM}: instead of projecting the output token onto a BPE embedding matrix, we project it onto multiple \textit{character} matrices in parallel. This allows us to enjoy the benefits of using a large vocabulary for tokenization, thereby using fewer tokens to represent each text, while substantially reducing the processing time of each output token.
  }

  \label{fig: arch_figure}
\end{figure}

We propose SpeLLM, a model that replaces token-level selection with \textit{simultaneous} character-level predictions. Our approach uses $k$ parallel linear heads over the final hidden state, each predicting a single character~(see~\cref{fig: arch_figure}). 
This design dramatically reduces the size of the output layer while combinatorially expanding the expressive capacity of the model. It also allows expanding the input vocabulary even further without increasing costs during output. We present a self-distillation approach, in which a baseline (BPE-based) LLM serves as a teacher, and a SpeLLM version of that model is the student.

We apply our method to four LLMs, and  show that their SpeLLM variants achieve competitive performance across multiple benchmarks, including GSM8K~\citep{GSM8K} and CNN/Daily Mail~\cite{CNN_daily_mail}. Additionally, these SpeLLMs offer an average speedup of 5.1\% across models. We further analyze SpeLLM’s performance as a function of the number of linear output heads, and find that it handles well various lengths of tokens.

Our contributions are as follows:  (1) We propose the SpeLLM architecture, which replaces the standard token-level output head with a sequence of character-level heads; (2) We present a self-distillation approach for adapting an existing LLM to this architecture; and (3) We empirically demonstrate that SpeLLM is faster than the teacher LLM, while maintaining comparable results in a wide range of tasks.

\section{SpeLLM: a spelling-based LLM} \label{sec:method}

We introduce SpeLLM---a novel LLM architecture, which replaces the BPE output head with a sequence of character heads, which decode in parallel. Below we describe the SpeLLM architecture~(\cref{sec:architecture}), the training procedure~(\cref{sec:training_procedure}), our loss functions~(\cref{sec:loss}), and our auto-correct procedure~(\cref{sec:autocorrect}).

\subsection{Architecture}\label{sec:architecture}

Consider an LLM $M$ with a BPE vocabulary of size $S$. Let $h \in \mathbb{R}^d$ be the hidden state corresponding to the current token at the final LLM layer. Consider also a character vocabulary of size $s$~($s << S$). The SpeLLM version of $M$~(denoted SpeLLM($M$)) maintains the same input vocabulary and model layers as $M$, but replaces the original output head with $k$ linear heads, each with $s$ rows~(denoted $L_i$). Each of the heads corresponds to an output character at a different position. To generate the next output token, SpeLLM($M$) makes a series of predictions \textit{in parallel}, where the spelling of the next word is
$c_i = \text{argmax}(\text{Softmax}(L_i(h)))$. 
The final string is constructed by concatenating $c_1,…,c_k$. To support for tokens shorter than $k$, we add a special padding symbol. See~\cref{fig: arch_figure} for illustration.

\subsection{BPE-to-SpeLLM self-distillation} \label{sec:training_procedure}
We introduce a self-distillation approach, where the teacher is the original (BPE-based) model ($M$), and the student is SpeLLM($M$). To minimize training costs, we only train the new linear layers, and  fine-tune the last five transformer feed-forward layers. This approach updates only a small, relatively shallow portion of the model, which reduces the cost of fine-tuning.
Our goal is for SpeLLM($M$) to match $M$'s BPE predictions. To do so, we run inference over a dataset and record $M$'s top-5 predictions, while ignoring the original gold word. We than compute the loss~(see below) with respect to each of the five tokens, and select the one with the lowest loss as our sole label.
This procedure is important in cases of model uncertainty, which might lead to the characters representing a blend of several different words. For example, given the input ``He is going to eat,'' the model might assign high probabilities to both ``\textcolor{ramenColor}{Donuts}''  and ``\textcolor{pizzaColor}{Pizza}''. Consequently, the character-level heads may produce an unintended string, e.g., ``\textcolor{ramenColor}{D}\textcolor{pizzaColor}{izza}''. To prevent such blending and encourage coherent spelling of individual tokens, we only take the loss with respect to the most similar word.

\subsection{Loss function}\label{sec:loss}

When designing our loss function, we want to maintain $M$'s token-level information. This is because solely relying on the character-level loss could degrade these representations, leading to undesirable phenomena such as words with similar spellings but different semantics collapsing into overly similar embeddings~(e.g., making the representation of ``book'' and ``look'' over similar). To mitigate this, we introduce an auxiliary token-level loss that regularizes the model to preserve its original token representations. During training, we retain and update the token prediction head to support this constraint. Overall, this leads us to formulate the loss as:\footnote{We also experimented with headless loss~\citep{headless_loss}. As it did not show significant benefits, we exclude it.
}

$$\mathcal{L} = \mathcal{L}_{\text{char}} + \mathcal{L}_{\text{token}}$$

\paragraph{Characters Loss}

Let $char\_logits_i$ denote the logits of the $i$th character head, and let $char\_probs_i$ denote $\text{Softmax}(char\_logits)$.

$$char\_argmax=[\text{argmax}(char\_probs_1),...,
\text{argmax}(char\_probs_k)]$$

From the precomputed top-3 token predictions at a given position, we select the candidate whose character sequence shares the largest number of position-wise matches with $char\_argmax$ and denote it with $similar$. Ties are broken by choosing the candidate with the highest original model probability.
The character loss is then:
$$\mathcal{L}_{\text{char}} = \sum_{i=0}^{k-1} \text{CrossEntropy}(char\_logits_i, similar_i)$$

\paragraph{Token Loss}

We denote by $top\_5$ the tensor with the top-5 probabilities  according to $M$ at the indexes of its top-5 tokens and zeros elsewhere. We denote by $token\_logits$ the logits obtained via the SpeLLM token head  for these 5 words.

The token loss is formulated as:

$$\mathcal{L}_{\text{token}} = \text{CrossEntropy}(token\_logits, top\_5)$$

\subsection{Auto-correction}\label{sec:autocorrect}

To further improve spelling performance, we employ an auto-correct mechanism: If the string we obtain matches any existing token, we output it. Otherwise, we filter all tokens in the BPE vocabulary where each character is in the top-3  in its corresponding SpeLLM position. We then apply a token-level linear head which consists of a vector representing only these tokens, and apply it over the $h$. We apply softmax and choose the token with the highest probability.

\section{Experimental setup}
\subsection{Base models}

We experiment with four open-source LLMs: Llama3.2-3B, Llama3-8B, Gemma2-2B, and Gemma2-9B \citep{gemma_2}. We append $k = 10$ linear layers with character-level vocabulary containing 105 symbols,\footnote{We note that a very small number of tokens are longer than 10 characters. In such cases, we expect the model to generate the first 10 characters, and the following token to generate the next. In practice, this happens quite rarely---for less than 3.24\% of the tokens in FineWeb-Edu. See~\cref{paragraph:token_coverage} for more details.} including the English alphabet~(lowercase and uppercase), digits, punctuation, and a padding symbol. It also includes a symbol for characters not in the vocabulary~(e.g, characters from non-latin scripts). Diacritical marks~(e.g., accents) are stripped from input during both training and evaluation.

\subsection{Training details}

All models are distilled using the first 500,000 samples\footnote{Samples which are too long to fit in the GPU are omitted.} from the \texttt{sample-10BT} subset of the FineWeb-Edu dataset \citep{FineWeb}, a high-quality corpus of educational web content. As explained in~\cref{sec:training_procedure}, we save the top-5 predictions and their probabilities for each position in the text for training. During both training and evaluation, we truncate the samples to a maximum of 1,400 tokens. Additional training details are provided in \cref{paragraph: additional_training_details}.

\subsection{Evaluation\label{section: benchmarks}}

\paragraph{Intrinsic evaluation}
We evaluate SpeLLM on 5,000 unseen samples from the FineWeb-Edu subset. Correctness is assessed by comparing SpeLLM’s output against the top-5 predictions of the model's token-level head. Our evaluation is based on three match criteria: (1) \textit{Full exact match} as a prediction that spells out every character of any one of the top-5 target tokens exactly; (2) \textit{10-character match} is defined as correctly spelling the first 10 characters of a token that exceeds 10 characters in length; (3) \textit{Match for prefix} refers to instances where the model generates a prefix of a top-5 token followed by padding. Although the predicted token is not in the top-5 list, it contains no spelling errors.  
We evaluate model performance both with and without the AutoCorrect mechanism~(\cref{sec:autocorrect}).

\paragraph{Downstream tasks}
We also experiment with several different downstream tasks: First, we use 500 samples from BoolQ~\citep{BoolQ} and ARC-Easy~\citep{ARC}, both multiple choice questions. We also use 500 samples from GSM8K~\citep{GSM8K} a dataset which consists of math questions, to study the effect of SpeLLM on the model's reasoning capabilities. Finally, we use 100 samples from CNN/Daily Mail~\cite{CNN_daily_mail}, a text summarization benchmark. In GSM8K, we use 6-shot for Llama3.2-3B, Llama3-8B, and Gemma2-2B and 3-shot for Gemma2-9B. Other benchmarks are used in zero-short settings. We evaluate results with and without AutoCorrect.

\section{Results \label{section:results}}

\begin{table}[!ht]

    \caption{Intrinsic evaluation of SpeLLM. \textit{Full exact math} rows indicate the accuracy of predicting any of teacher's top-5 words exactly; \textit{10-character match} indicate the accuracy of predicting the first 10 characters of any top-5 word; and \textit{Match for prefix} is the proportion of tokens where a prefix of one of the top-5 words is predicted; \textit{Total} is the sum of all three lines. SpeLLM matches ~(at least partially) 94.89\% of the tokens~(without AutoCorrect) and 97.57\% of the words~(with it). }
    \centering

\begin{tabular}{ccccccc}
\toprule
\textbf{\shortstack{Generation\\Type}} & \textbf{\shortstack{Match\\Type}} & 
\textbf{\shortstack{Llama3.2\\3B}} & \textbf{\shortstack{Llama3\\8B}} & 
\textbf{\shortstack{Gemma2\\2B}} & \textbf{\shortstack{Gemma2\\9B}} & 
\textbf{Average} \\
\midrule

\multirow{4}{*}{\shortstack{SpeLLM}} 
  & Full exact match       & 91.82 & 91.31 & 91.66 & 92.23 & 91.75 \\
  & 10-character match    &  \phantom{0}1.68 &  \phantom{0}1.72 &  \phantom{0}1.74 &  \phantom{0}2.03 &  \phantom{0}1.79 \\
  & Match for prefix       &  \phantom{0}1.42 &  \phantom{0}1.49 &  \phantom{0}1.28 &  \phantom{0}1.18 &  \phantom{0}1.34 \\
  & \textbf{Total}                  & 94.92 & 94.52 & 94.67 & 95.45 & 94.89 \\
\midrule
\multirow{4}{*}{\shortstack{SpeLLM\\+\\AutoCorrect}} 
  & Full exact match       & 95.17 & 94.57 & 94.84 & 95.15 & 94.93 \\
  & 10-character match    &  \phantom{0}1.68 &  \phantom{0}1.72 &  \phantom{0}1.74 &  \phantom{0}2.03 &  \phantom{0}1.79 \\
  & Match for prefix       &  \phantom{0}0.86 &  \phantom{0}0.94 &  \phantom{0}0.83 &  \phantom{0}0.75 &  \phantom{0}0.85 \\
  & \textbf{Total}                  & 97.71 & 97.23 & 97.41 & 97.93 & 97.57 \\

\bottomrule
\end{tabular}

    \label{table: SpeLLM Performance}
\end{table}

\paragraph{Intrinsic evaluation}
\Cref{table: SpeLLM Performance} presents our results. SpeLLM generates a full match in almost 92\% of the cases on average, and almost 95\% with AutoCorrect. Considering also prefix matches~(with no spelling errors) the numbers jump to 94.89\% and 97.57\%, respectively. 
We note that the prefix match is not trivial: across models, the average prefix length is 4.18 characters, while the average target string is 6.27 characters. See~\cref{prefix_len} (\cref{paragraph: prefixes}) for the distribution of prefix lengths across models.

\begin{figure}[ht]
  \centering

  \includegraphics[width= 1 \textwidth]{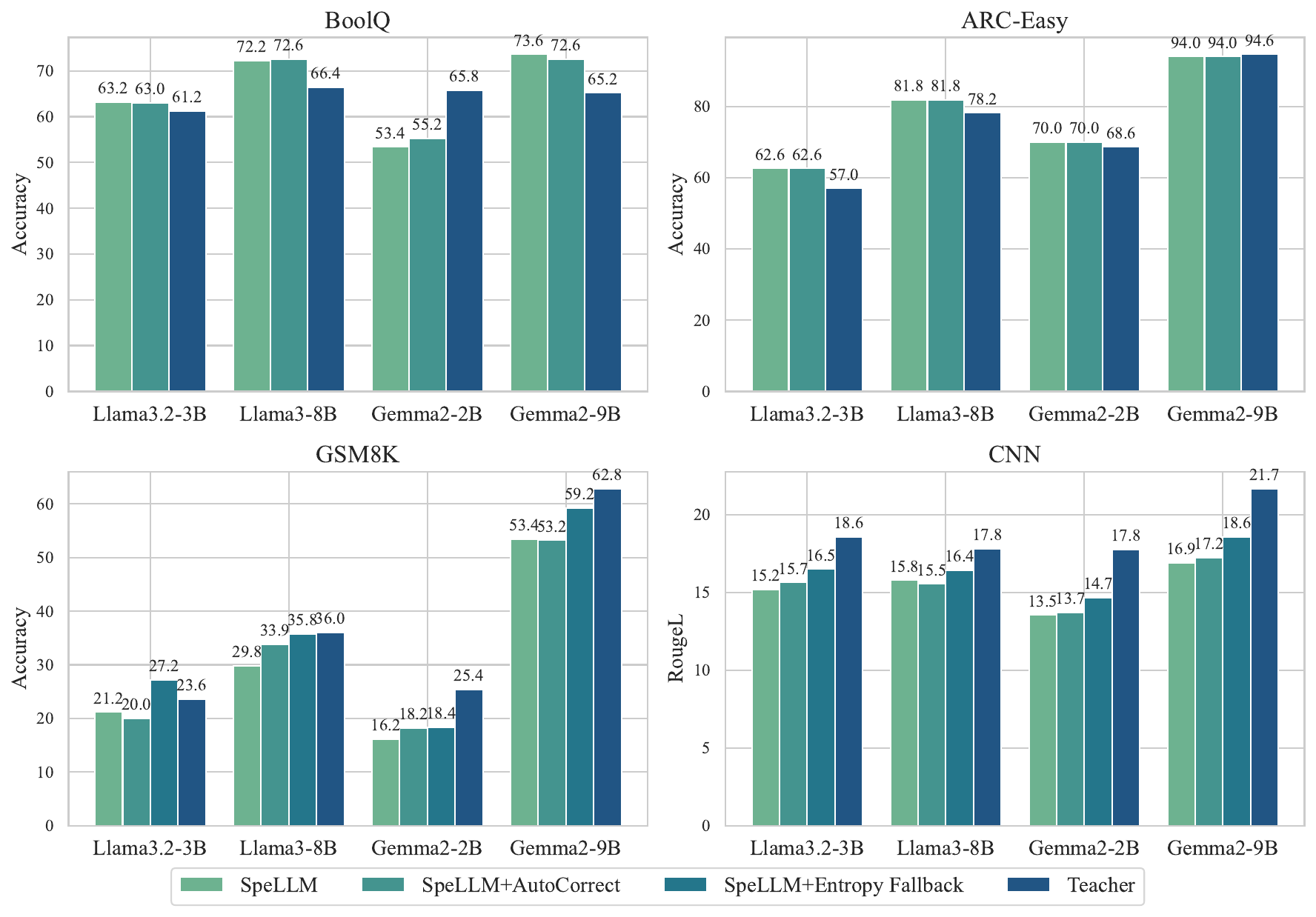}
  \caption{Results of SpeLLM on Downstream Tasks. SpeLLM is comparable or superior to the teacher model on BoolQ and ARC-Easy. On GSM8K and CNN/Daily Mail, there is a gap in favor of the teacher, which is partially closed using AutoCorrect and Entropy fallback~(see \cref{entropy_analysis}).}

  \label{fig:downstream_res_barplot}
\end{figure}

\paragraph{Downstream tasks}
Downstream results are shown in \cref{fig:downstream_res_barplot}. SpeLLM performance on BoolQ and ARC-Easy remains comparable to the teacher performance, and even surpasses it in many cases, often with a substantial margin of up to 6.2\%. An exception is Gemma2-2B on BoolQ, which suffers a substantial drop in performance. In contrast, a small degradation is observed on GSM8K and CNN/Daily Mail. Moreover, unlike the intrinsic evaluation, here we observe that in most cases, AutoCorrect doesn't improve performance substantially, and in some cases even hurts performance.

\paragraph{Runtime \label{runtime_section}}

\begin{figure}[ht]
  \centering

  \includegraphics[width= 1 \textwidth]{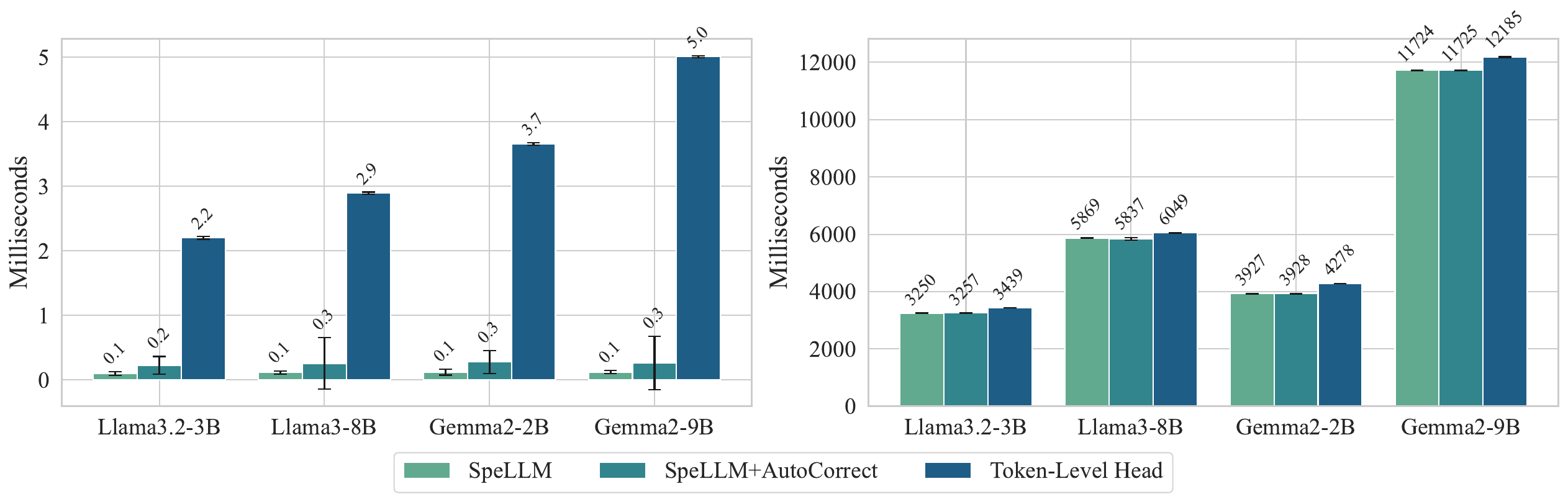}
  \caption{Runtime analysis of SpeLLM vs.~the teacher LLM across models. \textbf{Left}: SpeLLM dramatically reduces the cost of projecting the final hidden layer;  \textbf{Right}: when considering the runtime of the entire decoding process, the gaps are smaller, but still noticeable~(5.1\% runtime reduction on average). The addition of the AutoCorrect component is negligible.}

  \label{fig: all_run_speeds}
\end{figure}

We evaluate the runtime performance of SpeLLM relative to the teacher LLM under two conditions.
First, we isolate the output head by computing the final hidden states from the last model layer and measuring only the time taken by the output head, either SpeLLM or a standard linear projection. This is averaged across over 5,000 samples.
Second, we measure the generation latency by prompting the model with a 100-token input from FineWeb-Edu and timing the generation of the subsequent 99 tokens (excluding prompt processing time). This is done over 500 samples.

As shown in \cref{fig: all_run_speeds}, SpeLLM (with and without AutoCorrect) achieves an average speedup of 5.1\% over the standard token-level head, and the differences between them are neglibgble. When isolating the time for the decoding head, we observe a substantial difference between token-level decoding head and SpeLLM.

\section{Analysis}

\subsection{Performance with various number of character-heads}
Our results so far were obtained using 10 character heads. Here we aim to study the effect using a different number of heads. To do so, we perform self-distillation on Llama3.2-3B using configurations with $k=5,10,15$ output heads, and evaluate all models using our intrinsic evaluation~(\cref{table: SpeLLM Performance}). 

Our results(\cref{table: k_heads_results}) show that that the $k=15$ configuration achieves the highest exact match rates, while $k=5$ fails to recover many target words. Nonetheless, considering also partial matches, using fewer tokens leads to higher overall accuracy (without AutoCorrect) or similar accuracy (with it).

\begin{table}[h]

\caption{Character-level match accuracy of SpeLLM across varying numbers of output heads. As the number of heads increase, \textit{Exact match} scores increase. However, considering also the $k$ character match, the total numbers of all variants are similar, with 5 heads even outperforming the others in SpeLLM without AutoCorrect.}
\label{table: k_heads_results}
\vspace{5pt}
\centering

\begin{tabular}{llccc}
\toprule
\textbf{Text Generation} & \textbf{Match Type} &
\textbf{5 Heads} &
\textbf{10 Heads} &
\textbf{15 Heads}\\
\midrule
\multirow{3}{*}{\textbf{SpeLLM}} 
 & Exact match         & \phantom{0}72.39 & \phantom{0}91.82 & \phantom{0}\textbf{93.32} \\
 & $k$-character match & \phantom{0}22.22 & \phantom{00}1.68 & \phantom{00}0.00 \\
& \textbf{Total}               & \phantom{0}\textbf{94.61} & \phantom{0}93.50 & \phantom{0}93.32 \\
\midrule
\multirow{3}{*}{\begin{tabular}[c]{@{}l@{}}\textbf{SpeLLM +}\\\textbf{AutoCorrect}\end{tabular}} 
 & Exact match         & \phantom{0}74.59 & \phantom{0}95.17 & \phantom{0}\textbf{96.75} \\
 & $k$-character match & \phantom{0}22.22 & \phantom{00}1.68 & \phantom{00}0.00 \\
 & \textbf{Total}               & \phantom{0}96.81 & \phantom{0}\textbf{96.86} & \phantom{0}96.76 \\
\bottomrule
\end{tabular}

\end{table}

\begin{figure}[ht]
  \centering

  \includegraphics[width= 1 \textwidth]{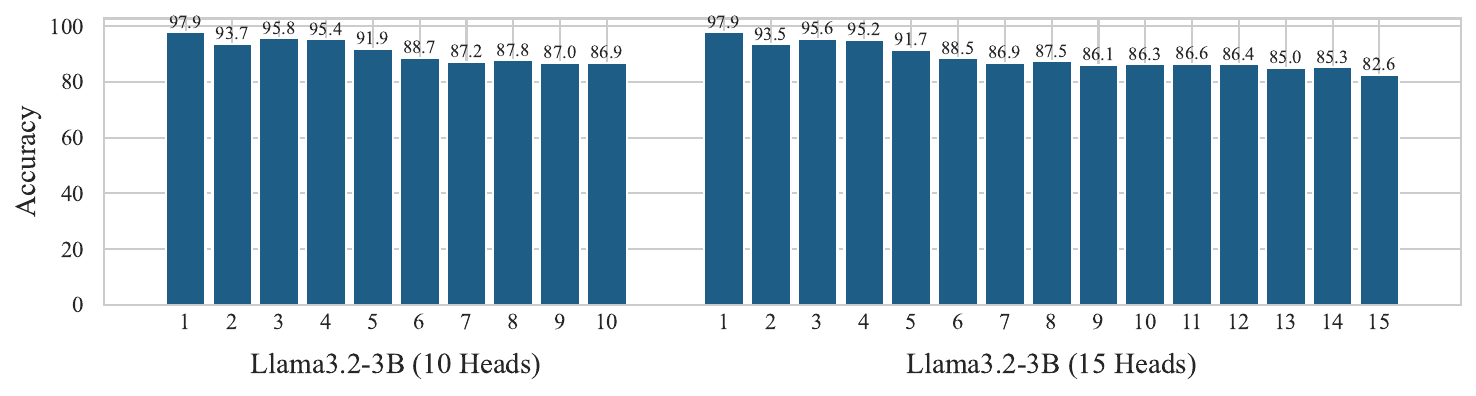}
  \caption{Accuracy vs.~length of the nearest token~(i.e., the token with the lowest number of mistakes compared to the SpeLLM argmax token), for both 10 and 15 heads. Accuracy values remain high~($>95\%$) for the first four characters, but start to decline later, though not very steeply.}

  \label{fig: len_vs_correctness_small}
\end{figure}

\subsection{Accuracy over various token lengths \label{paragraph: acc_vs_len}} We next aim to study SpeLLM's performance on tokens of different lengths. To do so, we select,
for each predicted token, the closest match from the top-5 teacher outputs based on the fewest spelling mistakes, with probability serving as a tie-breaker. We then group the predictions by the length of the best-matching string and compute accuracy within each group.

We evaluate the model using Llama3.2-3B variants with $k=10$ and $k=15$ linear output heads. As shown in \cref{fig: len_vs_correctness_small}, the models achieve significantly higher accuracy on shorter tokens, particularly those up to four characters long. However, beyond a token length of 7, accuracy plateaus and the degradation is relatively mild, even in the more complex $k=15$ head setting.

\subsection{Entropy analysis\label{entropy_analysis}} 

To gain deeper insight into when SpeLLM performs well or fails, we analyze the relationship between output entropy and prediction accuracy. This allows us to assess both the student model's behavior and its alignment with the pretrained teacher's uncertainty.

\begin{figure}[ht]
  \centering

  \includegraphics[width= 1 \textwidth]{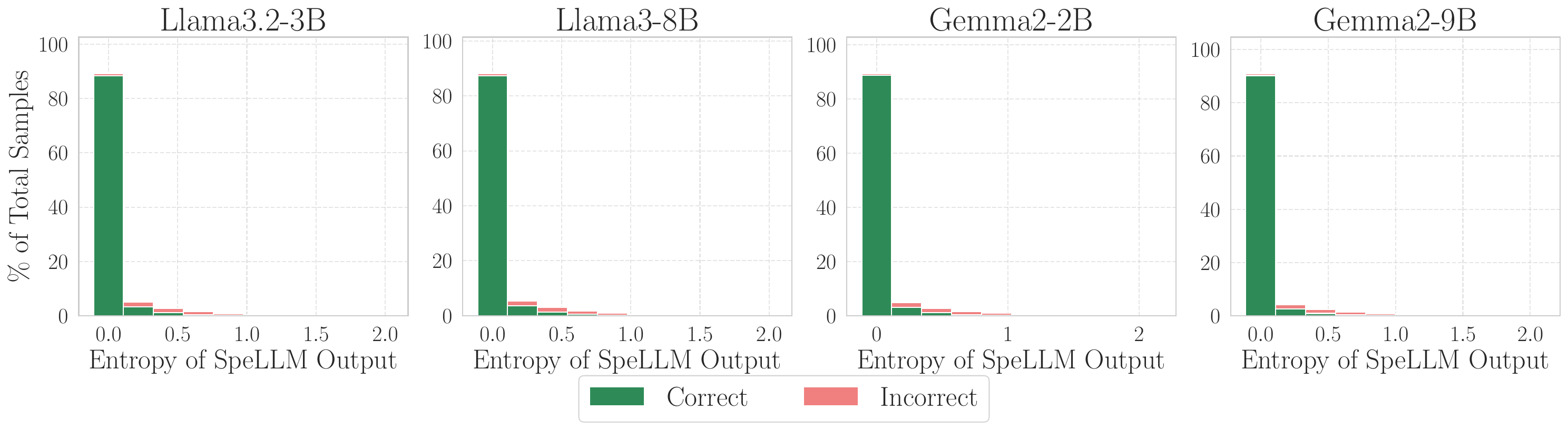}

\caption{SpeLLM predictions grouped by mean character-level entropy with accuracy indicated by color. The majority of samples show low entropy, with accuracy declining steeply as entropy rises.}

  \label{entropy_vs_correctness}
\end{figure}

\paragraph{SpeLLM entropy and prediction correctness} We compare output entropy to prediction correctness. We compute the entropy of each SpeLLM head, and report the average across heads. We split the results into eight equidistant bins. As shown in \cref{entropy_vs_correctness}, the model achieves an average accuracy of 99.16\% in the first bin. However, as output entropy increases, accuracy degrades sharply, dropping to 65.18\%  in the second bin. This suggests that entropy in SpeLLM’s output serves as a strong indicator of prediction accuracy.

Motivated by this observation, we propose an \textit{entropy-fallback} mechanism: when the mean output entropy across SpeLLM’s heads exceeds a predefined threshold, \footnote{We set the threshold at 0.22 for all models.} we fall back to using the full token-level head for prediction. We evaluate this mechanism on GSM8K and CNN/Daily Mail. As shown in \cref{fig:downstream_res_barplot}, entropy-fallback consistently outperforms AutoCorrect and yields results that closely approach those of the teacher model in GSM8K, even surpassing it with Llama-3.2-3B. On CNN/Daily Mail, results also improve, but are still below the teacher model.

\begin{figure}[ht]
  \centering

  \includegraphics[width= 1 \textwidth]{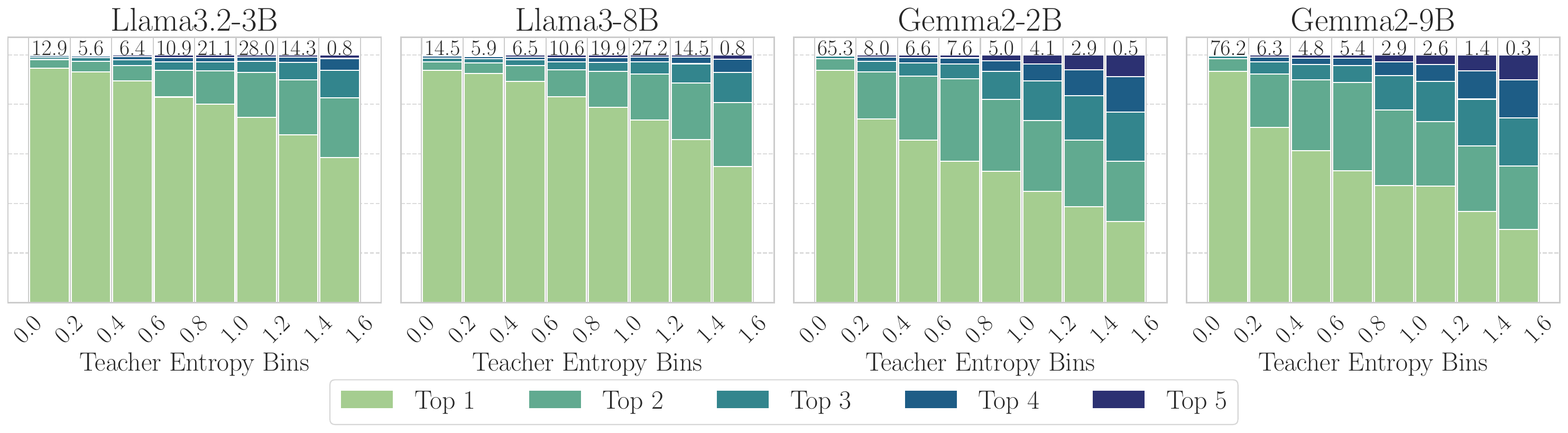}

    \caption{Relationship between teacher model entropy and SpeLLM's alignment with teacher predictions. When the teacher is confident (low entropy), SpeLLM tends to select the top-1 prediction. As entropy increases, SpeLLM increasingly selects lower-ranked teacher candidates. Percentages indicated above the bins correspond to the proportion of samples falling into each bin.}

  \label{top_1_pref}
\end{figure}

\paragraph{Top-k preference} We also investigate how SpeLLM allocates preference among the top-k predictions of the pretrained model, given that it is trained to predict any of the top-3 tokens. Specifically, we examine how the rank of the selected token  varies with the entropy of the pretrained model. As shown in \cref{top_1_pref}, SpeLLM overwhelmingly favors the top-1 token when the pretrained model exhibits low entropy, i.e., when it is confident in its prediction. As the pretrained model’s entropy increases, indicating higher uncertainty, SpeLLM becomes more likely to select lower-ranked predictions. 

\paragraph{Correlation between teacher and student entropy} To quantify the alignment between student and teacher uncertainty, we compare their respective entropy values. We compare SpeLLM’s mean output entropy with that of the corresponding pretrained model. We compute the Pearson correlation between the entropy of the pretrained model (approximated using the top-5 predicted token probabilities) and the mean output entropy of SpeLLM. For Gemma, we observe a strong correlation: 0.44 for Gemma2-2B and 0.43 for Gemma2-9B. In contrast, the correlation is substantially weaker for Llama models, with values of 0.08 for Llama-3.2-3B and 0.07 for Llama3-8B. This discrepancy is also reflected in the entropy distributions shown in \cref{top_1_pref}: Gemma’s outputs tend to exhibit low entropy, with over 65\% of samples falling in the lowest entropy bin, whereas Llama’s outputs are more uniformly spread across the entropy spectrum. This may partly account for the differing degrees of alignment between the pretrained and fine-tuned models.

\subsection{AutoCorrect analysis}

\begin{figure}[ht]
  \centering

  \includegraphics[width= 1 \textwidth]{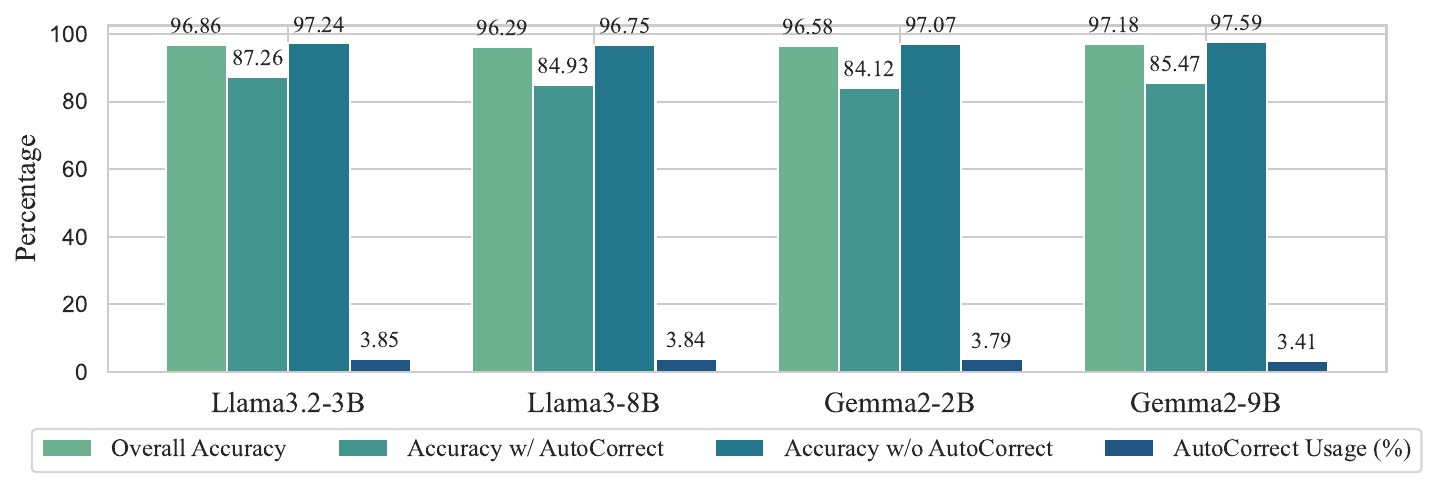}

  \caption{Analysis of AutoCorrect performance, including overall accuracy, accuracy when AutoCorrect is activated (i.e., the output string is not a valid token), and accuracy when the output string is a valid token. Valid token outputs are mostly correct, but accuracy decreases when correction is needed. AutoCorrect is triggered for 3.7\% of the time.}

  \label{fig: ac_analysis}
\end{figure}

\Cref{section:results} demonstrates that AutoCorrect considerably improves performance. We proceed by analyzing its usage frequency and effect on downstream performance.
As can be seen in \cref{fig: ac_analysis}, The AutoCorrect module is triggered only when the model outputs a string that does not match any existing token. Our analysis~(\cref{fig: ac_analysis}) shows that this happens in just 3.72\% of all predictions across models. We also observe a strong correlation between the need to invoke AutoCorrect and prediction errors: when the output is already a valid token (i.e., AutoCorrect is not invoked), the model is correct 97.16\% of the time. In contrast, when AutoCorrect is triggered, accuracy drops to 85.44\%.

Next, we look at the number of candidates in cases where AutoCorrect is applied, i.e., when
the output is not a valid token.

We find that, on average across models, there is a median of 31.7 candidates. In 0.3\% of applicable cases, no candidates are generated. In the top 0.5\% of cases, the average number of candidates rises to 7783.2, while the top 1\% see an average of 151.5 candidates.

\section{Related work}

\paragraph{Vocabulary size \& efficiency} Vocabulary size has long been recognized as a source of inefficiency in language models. 
\citet{vocab_reduction_for_SLM, t_free_tokenizer} have recognized that a significant proportion of parameters in small models are allocated to the embedding matrix. Meanwhile, \citet{fr_spec} have shown that the 25\% most common tokens are used more than 95\% of the time, and proposed frequency-based speculative decoding. \citet{vocab_compression, vocab_transfer, vocab_reduction_for_SLM} have proposed vocabulary reduction methods such as partial vocabulary loading. Others mitigated training inefficiencies through alternative approaches \citep{vocab_efficient_finetuning, headless_loss}.  \citet{t_free_tokenizer} employed a compressed decoding head for efficient text generation. We, instead, propose to use multiple small linear heads, which combined, are still notably smaller than the original linear head. The computational cost of the softmax layer, in particular, scales with vocabulary size and has motivated \citet{efficient_softmax} and subsequent works \cite{softermax_efficient_softmax, adaptive_efficient_softmax} to improve its efficiency.
In SpeLLM, the use of smaller linear output layers enables softmax computation over tensors that are less than 1\% the size of the original output space, substantially reducing runtime, as shown in \cref{runtime_section}.

\paragraph{Multiple linear heads in transformers} 
Recent work has investigated multi-token prediction using multiple linear heads. \citet{multi_token_pred} proposed generating multiple tokens by assigning a separate head to each token. Similarly, \citet{medusa} employed multiple heads to predict entire words simultaneously, incorporating tree attention as an alternative to speculative decoding. 
\citet{multi_lingual_multi_vocab} proposed using multiple linear heads, corresponding to different languages in multi-lingual LLMs. We also propose using multiple linear heads, such that the $k$ head predicts the $k$th character in the output string.

\paragraph{Alternative transformer variants}
Prior work has explored pre-training models directly on byte-level \cite{byt5} or character-level sequences \cite{canine}. Namely, MEGABYTE~\citep{mega_byte} and BLT~\citep{meta_blt} train lightweight local decoder-only transformers over a larger global model to predict bytes auto-regressively. In contrast, our approach is much simpler: it uses self-distillation to convert any existing pre-trained model to predict characters instead of tokens, with minimal changes. Furthermore, our decoding operates in parallel using $k$ independent linear character heads, predicting entire strings at a time.

\section{Limitations and broader impact}

\paragraph{Limitations}

Despite decoupling output representations, our method still relies on token-level encoding both for training and AutoCorrect, which constrains full departure from BPE modeling. Additionally, despite achieving faster decoding, the majority of computational cost remains in the transformer layers, constraining overall efficiency gains. Finally, while faster and simpler, our approach is not as accurate on as the teacher model in some tasks (GSM8K and CNN/Daily Mail).

\paragraph{Broader Impact}

This work presents a step toward more efficient and accessible language models by enabling faster generation. Across models, we observe a 5.1\% speedup in decoding runtime. More importantly, our approach offers a potential pathway to reducing the vocabulary size bottleneck, which disproportionately affects low-resource languages due to their poor token coverage in conventional models~\citep{ahia-etal-2023-languages,sengupta2023jais, petrov2024language}. By representing outputs as flexible character-level strings, our method can reduce dependency on large token vocabularies and make model deployment more cost-effective. This could democratize access to high-quality LLMs in underserved linguistic communities. 

\section{Conclusions}
We introduced SpeLLM---a novel architecture that predicts output tokens character by character using $k$ parallel linear heads. This design significantly increases the potential output space while reducing computational costs in comparison with traditional token-based models.
We have also shown how existing pre-trained models can be adapted to SpeLLM through a self-distillation approach. Our empirical results indicate that SpeLLM decodes 5.1\% faster than conventional single-head models, while producing comparable results.
Our results pave the way towards using far larger LLM vocabularies, which would allow for increased support for underrepresented languages. They also have the potential to support the generation of multiple tokens in one character sequence, which could further reduce sequence lengths, and decoding costs accordingly.

\bibliography{main.bib}

\begin{thebibliography}{32}
\providecommand{\natexlab}[1]{#1}
\providecommand{\url}[1]{\texttt{#1}}
\expandafter\ifx\csname urlstyle\endcsname\relax
  \providecommand{\doi}[1]{doi: #1}\else
  \providecommand{\doi}{doi: \begingroup \urlstyle{rm}\Url}\fi

\bibitem[Gage(1994)]{bpe1}
Philip Gage.
\newblock {A new algorithm for data compression}.
\newblock \emph{C Users J.}, 12\penalty0 (2):\penalty0 23--38, February 1994.
\newblock ISSN 0898-9788.
\newblock \doi{10.5555/177910.177914}.

\bibitem[Sennrich et~al.(2015)Sennrich, Haddow, and Birch]{bpe2}
Rico Sennrich, Barry Haddow, and Alexandra Birch.
\newblock {Neural Machine Translation of Rare Words with Subword Units}.
\newblock \emph{arXiv}, August 2015.
\newblock \doi{10.48550/arXiv.1508.07909}.

\bibitem[Grattafiori et~al.(2024)Grattafiori, Dubey, Jauhri, Pandey, et~al.]{llama_3}
Aaron Grattafiori, Abhimanyu Dubey, Abhinav Jauhri, Abhinav Pandey, et~al.
\newblock {The Llama 3 Herd of Models}.
\newblock \emph{arXiv}, July 2024.
\newblock \doi{10.48550/arXiv.2407.21783}.

\bibitem[Team et~al.(2024)Team, Riviere, Pathak, Sessa, et~al.]{gemma_2}
Gemma Team, Morgane Riviere, Shreya Pathak, Pier~Giuseppe Sessa, et~al.
\newblock {Gemma 2: Improving Open Language Models at a Practical Size}.
\newblock \emph{arXiv}, July 2024.
\newblock \doi{10.48550/arXiv.2408.00118}.

\bibitem[Ahia et~al.(2023)Ahia, Kumar, Gonen, Kasai, Mortensen, Smith, and Tsvetkov]{ahia-etal-2023-languages}
Orevaoghene Ahia, Sachin Kumar, Hila Gonen, Jungo Kasai, David Mortensen, Noah Smith, and Yulia Tsvetkov.
\newblock Do all languages cost the same? tokenization in the era of commercial language models.
\newblock In Houda Bouamor, Juan Pino, and Kalika Bali, editors, \emph{Proceedings of the 2023 Conference on Empirical Methods in Natural Language Processing}, pages 9904--9923, Singapore, December 2023. Association for Computational Linguistics.
\newblock \doi{10.18653/v1/2023.emnlp-main.614}.
\newblock URL \url{https://aclanthology.org/2023.emnlp-main.614}.

\bibitem[Sengupta et~al.(2023)Sengupta, Sahu, Jia, Katipomu, Li, Koto, Marshall, Gosal, Liu, Chen, Afzal, Kamboj, Pandit, Pal, Pradhan, Mujahid, Baali, Han, Bsharat, Aji, Shen, Liu, Vassilieva, Hestness, Hock, Feldman, Lee, Jackson, Ren, Nakov, Baldwin, and Xing]{sengupta2023jais}
Neha Sengupta, Sunil~Kumar Sahu, Bokang Jia, Satheesh Katipomu, Haonan Li, Fajri Koto, William Marshall, Gurpreet Gosal, Cynthia Liu, Zhiming Chen, Osama~Mohammed Afzal, Samta Kamboj, Onkar Pandit, Rahul Pal, Lalit Pradhan, Zain~Muhammad Mujahid, Massa Baali, Xudong Han, Sondos~Mahmoud Bsharat, Alham~Fikri Aji, Zhiqiang Shen, Zhengzhong Liu, Natalia Vassilieva, Joel Hestness, Andy Hock, Andrew Feldman, Jonathan Lee, Andrew Jackson, Hector~Xuguang Ren, Preslav Nakov, Timothy Baldwin, and Eric Xing.
\newblock Jais and jais-chat: Arabic-centric foundation and instruction-tuned open generative large language models, 2023.
\newblock URL \url{https://arxiv.org/abs/2308.16149}.
\newblock {arXiv}:2308.16149.

\bibitem[Petrov et~al.(2023)Petrov, La~Malfa, Torr, and Bibi]{petrov2024language}
Aleksandar Petrov, Emanuele La~Malfa, Philip Torr, and Adel Bibi.
\newblock Language model tokenizers introduce unfairness between languages.
\newblock In \emph{Proc. of NeurIPS}, 2023.

\bibitem[Yu et~al.(2023)Yu, Simig, Flaherty, Aghajanyan, et~al.]{mega_byte}
Lili Yu, D{\ifmmode\acute{a}\else\'{a}\fi}niel Simig, Colin Flaherty, Armen Aghajanyan, et~al.
\newblock {MEGABYTE: Predicting Million-byte Sequences with Multiscale Transformers}.
\newblock \emph{arXiv}, May 2023.
\newblock \doi{10.48550/arXiv.2305.07185}.

\bibitem[Pagnoni et~al.(2024)Pagnoni, Pasunuru, Rodriguez, Nguyen, Muller, Li, Zhou, Yu, Weston, Zettlemoyer, Ghosh, Lewis, Holtzman, and Iyer]{meta_blt}
Artidoro Pagnoni, Ram Pasunuru, Pedro Rodriguez, John Nguyen, Benjamin Muller, Margaret Li, Chunting Zhou, Lili Yu, Jason Weston, Luke Zettlemoyer, Gargi Ghosh, Mike Lewis, Ari Holtzman, and Srinivasan Iyer.
\newblock {Byte Latent Transformer: Patches Scale Better Than Tokens}.
\newblock \emph{arXiv}, December 2024.
\newblock \doi{10.48550/arXiv.2412.09871}.

\bibitem[Itzhak and Levy(2021)]{spelling_bee}
Itay Itzhak and Omer Levy.
\newblock {Models In a Spelling Bee: Language Models Implicitly Learn the Character Composition of Tokens}.
\newblock \emph{arXiv}, August 2021.
\newblock \doi{10.48550/arXiv.2108.11193}.

\bibitem[Kaushal and Mahowald(2022)]{what_do_tokens_know}
Ayush Kaushal and Kyle Mahowald.
\newblock What do tokens know about their characters and how do they know it?
\newblock \emph{ArXiv}, abs/2206.02608, 2022.
\newblock URL \url{https://api.semanticscholar.org/CorpusID:249394509}.

\bibitem[Cobbe et~al.(2021)Cobbe, Kosaraju, Bavarian, Chen, Jun, Kaiser, Plappert, Tworek, Hilton, Nakano, Hesse, and Schulman]{GSM8K}
Karl Cobbe, Vineet Kosaraju, Mohammad Bavarian, Mark Chen, Heewoo Jun, Lukasz Kaiser, Matthias Plappert, Jerry Tworek, Jacob Hilton, Reiichiro Nakano, Christopher Hesse, and John Schulman.
\newblock {Training Verifiers to Solve Math Word Problems}.
\newblock \emph{arXiv}, October 2021.
\newblock \doi{10.48550/arXiv.2110.14168}.

\bibitem[Hermann et~al.(2015)Hermann, Ko{\ifmmode\check{c}\else\v{c}\fi}isk{\ifmmode\acute{y}\else\'{y}\fi}, Grefenstette, Espeholt, Kay, Suleyman, and Blunsom]{CNN_daily_mail}
Karl~Moritz Hermann, Tom{\ifmmode\acute{a}\else\'{a}\fi}{\ifmmode\check{s}\else\v{s}\fi} Ko{\ifmmode\check{c}\else\v{c}\fi}isk{\ifmmode\acute{y}\else\'{y}\fi}, Edward Grefenstette, Lasse Espeholt, Will Kay, Mustafa Suleyman, and Phil Blunsom.
\newblock {Teaching Machines to Read and Comprehend}.
\newblock \emph{arXiv}, June 2015.
\newblock \doi{10.48550/arXiv.1506.03340}.

\bibitem[Godey et~al.(2023)Godey, de~la Clergerie, and Sagot]{headless_loss}
Nathan Godey, {\ifmmode\acute{E}\else\'{E}\fi}ric de~la Clergerie, and Beno{\ifmmode\hat{\imath}\else\^{\i}\fi}t Sagot.
\newblock {Headless Language Models: Learning without Predicting with Contrastive Weight Tying}.
\newblock \emph{arXiv}, September 2023.
\newblock \doi{10.48550/arXiv.2309.08351}.

\bibitem[Penedo et~al.(2024)Penedo, Kydl{\ifmmode\acute{\imath}\else\'{\i}\fi}{\ifmmode\check{c}\else\v{c}\fi}ek, Allal, Lozhkov, Mitchell, Raffel, Von~Werra, and Wolf]{FineWeb}
Guilherme Penedo, Hynek Kydl{\ifmmode\acute{\imath}\else\'{\i}\fi}{\ifmmode\check{c}\else\v{c}\fi}ek, Loubna~Ben Allal, Anton Lozhkov, Margaret Mitchell, Colin Raffel, Leandro Von~Werra, and Thomas Wolf.
\newblock {The FineWeb Datasets: Decanting the Web for the Finest Text Data at Scale}.
\newblock \emph{arXiv}, June 2024.
\newblock \doi{10.48550/arXiv.2406.17557}.

\bibitem[Clark et~al.(2019)Clark, Lee, Chang, Kwiatkowski, Collins, and Toutanova]{BoolQ}
Christopher Clark, Kenton Lee, Ming-Wei Chang, Tom Kwiatkowski, Michael Collins, and Kristina Toutanova.
\newblock {BoolQ: Exploring the Surprising Difficulty of Natural Yes/No Questions}.
\newblock \emph{arXiv}, May 2019.
\newblock \doi{10.48550/arXiv.1905.10044}.

\bibitem[Clark et~al.(2018)Clark, Cowhey, Etzioni, Khot, Sabharwal, Schoenick, and Tafjord]{ARC}
Peter Clark, Isaac Cowhey, Oren Etzioni, Tushar Khot, Ashish Sabharwal, Carissa Schoenick, and Oyvind Tafjord.
\newblock {Think you have Solved Question Answering? Try ARC, the AI2 Reasoning Challenge}.
\newblock \emph{arXiv}, March 2018.
\newblock \doi{10.48550/arXiv.1803.05457}.

\bibitem[Nozaki et~al.(2025)Nozaki, Nakashima, Sato, Asaba, and Kawamura]{vocab_reduction_for_SLM}
Yuta Nozaki, Dai Nakashima, Ryo Sato, Naoki Asaba, and Shintaro Kawamura.
\newblock Efficient vocabulary reduction for small language models.
\newblock In Owen Rambow, Leo Wanner, Marianna Apidianaki, Hend Al-Khalifa, Barbara~Di Eugenio, Steven Schockaert, Kareem Darwish, and Apoorv Agarwal, editors, \emph{Proceedings of the 31st International Conference on Computational Linguistics: Industry Track}, pages 771--783, Abu Dhabi, UAE, January 2025. Association for Computational Linguistics.
\newblock URL \url{https://aclanthology.org/2025.coling-industry.64/}.

\bibitem[Deiseroth et~al.(2024)Deiseroth, Brack, Schramowski, Kersting, and Weinbach]{t_free_tokenizer}
Bj{\ifmmode\ddot{o}\else\"{o}\fi}rn Deiseroth, Manuel Brack, Patrick Schramowski, Kristian Kersting, and Samuel Weinbach.
\newblock {T-FREE: Subword Tokenizer-Free Generative LLMs via Sparse Representations for Memory-Efficient Embeddings}.
\newblock \emph{arXiv}, June 2024.
\newblock \doi{10.48550/arXiv.2406.19223}.

\bibitem[Zhao et~al.(2025)Zhao, Pan, Han, Zhang, Sun, Huang, Zhang, Zhao, Li, Wang, Liu, and Sun]{fr_spec}
Weilin Zhao, Tengyu Pan, Xu~Han, Yudi Zhang, Ao~Sun, Yuxiang Huang, Kaihuo Zhang, Weilun Zhao, Yuxuan Li, Jianyong Wang, Zhiyuan Liu, and Maosong Sun.
\newblock {FR-Spec: Accelerating Large-Vocabulary Language Models via Frequency-Ranked Speculative Sampling}.
\newblock \emph{arXiv}, February 2025.
\newblock \doi{10.48550/arXiv.2502.14856}.

\bibitem[Vennam et~al.(2024)Vennam, Joishy, and Kumaraguru]{vocab_compression}
Sreeram Vennam, Anish Joishy, and Ponnurangam Kumaraguru.
\newblock {LLM Vocabulary Compression for Low-Compute Environments}.
\newblock \emph{arXiv}, November 2024.
\newblock \doi{10.48550/arXiv.2411.06371}.

\bibitem[Gee et~al.(2024)Gee, Zugarini, Rigutini, and Torroni]{vocab_transfer}
Leonidas Gee, Andrea Zugarini, Leonardo Rigutini, and Paolo Torroni.
\newblock {Fast Vocabulary Transfer for Language Model Compression}.
\newblock \emph{arXiv}, February 2024.
\newblock \doi{10.18653/v1/2022.emnlp-industry.41}.

\bibitem[Williams and Aletras(2023)]{vocab_efficient_finetuning}
Miles Williams and Nikolaos Aletras.
\newblock {Vocabulary-level Memory Efficiency for Language Model Fine-tuning}.
\newblock \emph{arXiv}, September 2023.
\newblock \doi{10.48550/arXiv.2309.08708}.

\bibitem[Grave et~al.(2016)Grave, Joulin, Ciss{\ifmmode\acute{e}\else\'{e}\fi}, Grangier, and J{\ifmmode\acute{e}\else\'{e}\fi}gou]{efficient_softmax}
Edouard Grave, Armand Joulin, Moustapha Ciss{\ifmmode\acute{e}\else\'{e}\fi}, David Grangier, and Herv{\ifmmode\acute{e}\else\'{e}\fi} J{\ifmmode\acute{e}\else\'{e}\fi}gou.
\newblock {Efficient softmax approximation for GPUs}.
\newblock \emph{arXiv}, September 2016.
\newblock \doi{10.48550/arXiv.1609.04309}.

\bibitem[Stevens et~al.(2021)Stevens, Venkatesan, Dai, Khailany, and Raghunathan]{softermax_efficient_softmax}
Jacob~R. Stevens, Rangharajan Venkatesan, Steve Dai, Brucek Khailany, and Anand Raghunathan.
\newblock {Softermax: Hardware/Software Co-Design of an Efficient Softmax for Transformers}.
\newblock \emph{arXiv}, March 2021.
\newblock \doi{10.48550/arXiv.2103.09301}.

\bibitem[Baharav et~al.(2024)Baharav, Kang, Sullivan, Tiwari, Luxenberg, Tse, and Pilanci]{adaptive_efficient_softmax}
Tavor~Z. Baharav, Ryan Kang, Colin Sullivan, Mo~Tiwari, Eric Luxenberg, David Tse, and Mert Pilanci.
\newblock {Adaptive Sampling for Efficient Softmax Approximation}.
\newblock \emph{Advances in Neural Information Processing Systems}, 37:\penalty0 117580--117613, December 2024.
\newblock URL \url{https://proceedings.neurips.cc/paper_files/paper/2024/hash/d52dbd66219dc4e432e0bd4f9c25c4c3-Abstract-Conference.html}.

\bibitem[Gloeckle et~al.(2024)Gloeckle, Idrissi, Rozi{\ifmmode\grave{e}\else\`{e}\fi}re, Lopez-Paz, et~al.]{multi_token_pred}
Fabian Gloeckle, Badr~Youbi Idrissi, Baptiste Rozi{\ifmmode\grave{e}\else\`{e}\fi}re, David Lopez-Paz, et~al.
\newblock {Better {\&} Faster Large Language Models via Multi-token Prediction}.
\newblock \emph{arXiv}, April 2024.
\newblock \doi{10.48550/arXiv.2404.19737}.

\bibitem[Cai et~al.(2024)Cai, Li, Geng, Peng, et~al.]{medusa}
Tianle Cai, Yuhong Li, Zhengyang Geng, Hongwu Peng, et~al.
\newblock {Medusa: Simple LLM Inference Acceleration Framework with Multiple Decoding Heads}.
\newblock \emph{arXiv}, January 2024.
\newblock \doi{10.48550/arXiv.2401.10774}.

\bibitem[Chung et~al.(2020)Chung, Garrette, Tan, and Riesa]{multi_lingual_multi_vocab}
Hyung~Won Chung, Dan Garrette, Kiat~Chuan Tan, and Jason Riesa.
\newblock {Improving Multilingual Models with Language-Clustered Vocabularies}.
\newblock \emph{arXiv}, October 2020.
\newblock \doi{10.48550/arXiv.2010.12777}.

\bibitem[Xue et~al.(2021)Xue, Barua, Constant, Al-Rfou, Narang, Kale, Roberts, and Raffel]{byt5}
Linting Xue, Aditya Barua, Noah Constant, Rami Al-Rfou, Sharan Narang, Mihir Kale, Adam Roberts, and Colin Raffel.
\newblock {ByT5: Towards a token-free future with pre-trained byte-to-byte models}.
\newblock \emph{arXiv}, May 2021.
\newblock \doi{10.48550/arXiv.2105.13626}.

\bibitem[Clark et~al.(2021)Clark, Garrette, Turc, and Wieting]{canine}
Jonathan~H. Clark, Dan Garrette, Iulia Turc, and John Wieting.
\newblock {CANINE: Pre-training an Efficient Tokenization-Free Encoder for Language Representation}.
\newblock \emph{arXiv}, March 2021.
\newblock \doi{10.1162/tacl_a_00448}.

\bibitem[Loshchilov and Hutter(2017)]{adamw}
Ilya Loshchilov and Frank Hutter.
\newblock {Decoupled Weight Decay Regularization}.
\newblock \emph{arXiv}, November 2017.
\newblock \doi{10.48550/arXiv.1711.05101}.

\end{thebibliography}
\bibliographystyle{unsrtnat}

\appendix

\section{Appendix}

\paragraph{Accuracy over various token lengths} In \cref{paragraph: acc_vs_len}, we analyze the accuracy of models with varying number of character-level heads across different token lengths. Here we create a plot similar to \cref{fig: len_vs_correctness_small} with additional models (all with $k=10$ character level heads).

\begin{figure}[ht]
  \centering

    \includegraphics[width= 1\textwidth]{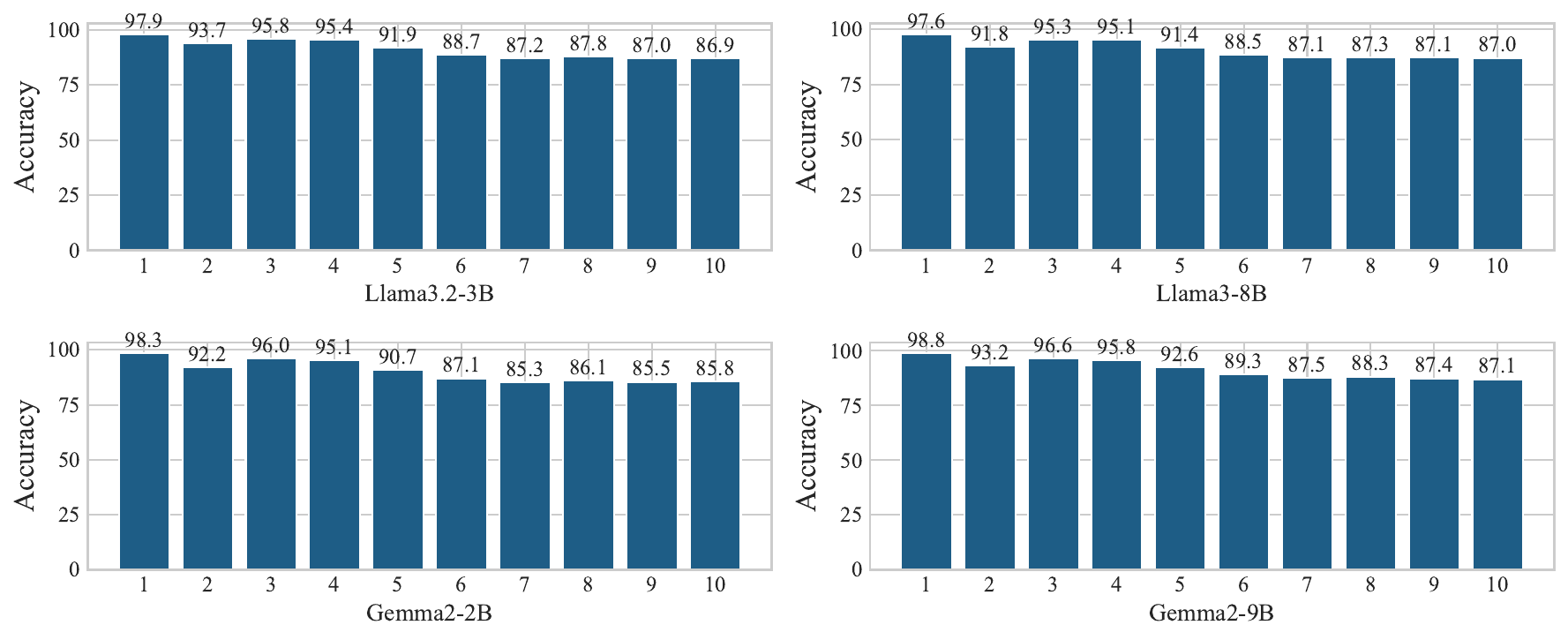}
  \caption{Accuracy vs.~length of nearest token across models (i.e. token with least number of mistakes compares
to argmax). }

  \label{fig: len_vs_correctness_full}
\end{figure}

\paragraph{Prefixes \label{paragraph: prefixes}} In \cref{section: benchmarks}, we note that in some cases, the model generates a prefix of one of the top-5 tokens followed by padding, without any spelling errors, instead of predicting a complete word. To highlight that this behavior is non-trivial, we present the average lengths of the generated prefix and the target word across models in such cases.

\paragraph{Additional Training Details \label{paragraph: additional_training_details}}
We adopt FP16 precision for training and use the AdamW optimizer \cite{adamw} with a learning rate of 5e-5 and a weight decay of 0.01. All models are trained on a single GPU: smaller models on Nvidia L40S, and larger models on Nvidia A100. 

\paragraph{Token Coverage\label{paragraph:token_coverage}}  Llama uses a vocabulary of 128K tokens, while Gemma models use 256K. Our approach employs 10 linear heads, each predicting 105 characters, for a total of 1,050 vectors, less than 0.85\% the size of either baseline's vocabulary.
Despite the small size of each head, we can represent 91\% of tokens composed exclusively of Latin characters. The remaining unrepresented tokens account for only 3.24\% of all tokens in FineWeb-Edu.

\begin{figure}[ht]
  \centering

  \includegraphics[width=1 \textwidth]{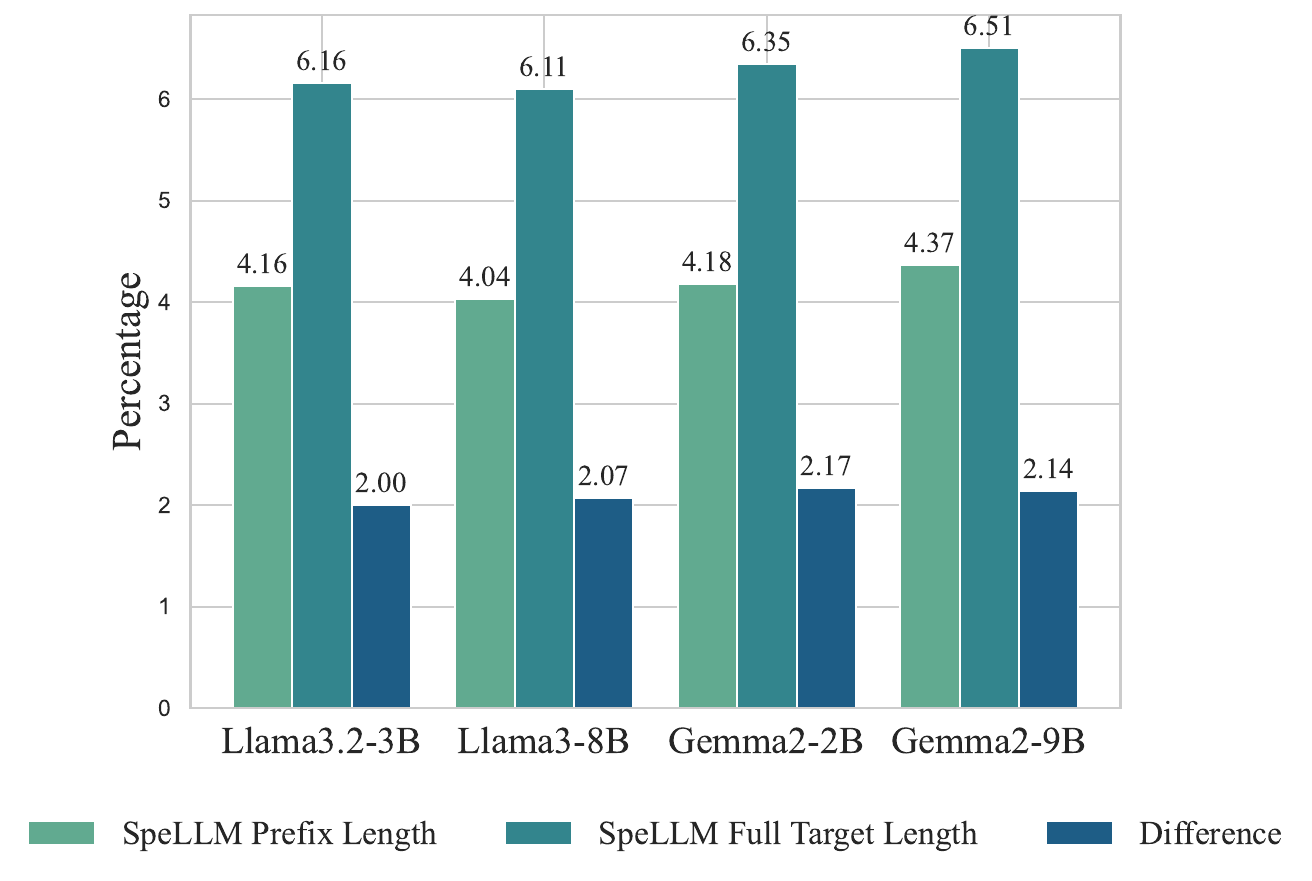}
  \caption{Mean length of the SpeLLM prefix and the corresponding target tokens.  
  }

  \label{prefix_len}
\end{figure}

\newpage

\end{document}